\documentclass[preprint,12pt]{elsarticle}

\usepackage{amsmath,amssymb,bm}
\usepackage{amsthm}
\usepackage{tikz}
\usetikzlibrary{calc}
\usetikzlibrary{positioning}
\theoremstyle{definition}

\theoremstyle{plain}
\newtheorem{proposition}{Proposition}

\journal{Journal of Mathematical Psychology}

\begin{document}

\begin{frontmatter}

\title{Contextual Information Allocation in Shared-State Cognitive Models: An Information-Theoretic Bound}

\author[inst1]{Song-Ju Kim}
\ead{kim@sobin.org}
\affiliation[inst1]{organization={SOBIN Institute LLC},
            city={Kawanishi},
            state={Hyogo},
            country={Japan}}

\begin{abstract}
Context-sensitive behavior can be modeled by enriching an internal state, by allowing a response
rule to access context directly, or by preserving a shared state while introducing an auxiliary
criterion or control variable. This paper isolates an information-theoretic constraint on the third
architecture. Let $C$ denote context, $S$ a candidate internal or latent state, $O$ an observable
response, and $M$ an auxiliary variable such that $O\perp C\mid(S,M)$. Then
\[
I(C;O\mid S)\le I(C;M\mid S)\le H(M\mid S).
\]
Once a shared state has been specified, residual context dependence in behavior therefore
lower-bounds both the context information and the conditional entropy that an auxiliary
context-mediating mechanism must carry. The bound is representation-relative rather than a
measure of state-space size or a universal contextuality measure. A worked recognition-memory
example shows how the quantity can be computed for payoff-induced criterion shifts and compared
across alternative representational allocations. Ontological contextuality and quantum probability
are treated as a separate specialization rather than as prerequisites for the cognitive interpretation.
More broadly, the framework provides a basis for analyzing context--memory--control trade-offs
in cognitive models and artificial agents that must maintain coherent behavior across changing
contexts under bounded internal representations.
\end{abstract}


\end{frontmatter}

\section{Introduction}
\label{sec:introduction}

Where should an intelligent system represent the information it needs to behave appropriately
across changing contexts? Should context-relevant distinctions be absorbed into a persistent
internal state, supplied directly by the current context, or carried by a separate memory or control
mechanism? This question spans both human and artificial intelligence. In cognition, the same
stimulus, memory cue, option, or question can produce different responses when instructions,
payoffs, preceding events, or other contextual conditions change. In contemporary artificial
agents, long-horizon interaction likewise requires mechanisms for retaining, retrieving, and
controlling task-relevant context; recent work treats memory as a key architectural component of
LLM-based agents \cite{ZhangAgentMemory2025}. A related recurrent context-switching study
makes the same allocation problem explicit computationally by comparing direct context access,
recurrent memory growth, and intervention on a shared latent state \cite{KimContextualControl2026}.

A substantial literature has addressed the complexity and sufficiency of cognitive state
representations. Discrete-state learning models can require a minimum number of states for
particular behavioral effects \cite{KatsikopoulosFisher2001}; Markov, partially observable, and
predictive-state approaches ask what information a state must retain for prediction and control
\cite{Puterman1994,Kaelbling1998,Littman2002}; and model-selection work has emphasized
complexity beyond parameter count \cite{Myung2000,Ashby2023}. Resource-rational approaches
likewise emphasize how limited representational and computational resources are allocated
\cite{Simon,LiederGriffiths2020}. These perspectives motivate a related question: after a
particular state representation has been chosen, how much context-sensitive behavioral
information remains outside that state?

There are several architectural answers. A model can internalize context-relevant distinctions in
a richer state, keep a shared state while allowing the response rule to access context directly, or
preserve both a shared state and a common response mechanism while introducing an auxiliary
criterion, task set, control state, retrieved context, or routing variable. These alternatives can
generate the same observable behavior while making different claims about where context
information resides.

The present paper develops an information-theoretic diagnostic for the third architecture. Let
$C$ denote context, $S$ a candidate internal or latent state, $O$ the observable response, and $M$
an auxiliary variable satisfying $O\perp C\mid(S,M)$. The main result implies
\[
I(C;O\mid S)\le I(C;M\mid S)\le H(M\mid S).
\]
Thus, after a candidate shared state has been fixed, the residual dependence of the response on
context places a lower bound on what an auxiliary mediator must carry. The quantity
$I(C;O\mid S)$ is representation-relative: it measures neither raw state-space size nor an
absolute property of the cognitive system, but the context dependence left unresolved by the
chosen state representation.

To make the use of the bound concrete, we develop a worked example based on decision-criterion
shifts in recognition memory. Signal-detection theory distinguishes evidence from decision
criterion \cite{MacmillanCreelman2005}, and payoffs and prior probabilities are known to alter
the cutoff for old--new responding \cite{HealyKubovy1978}. At a fixed candidate evidence state,
we calculate $I(C;O\mid S=s_0)$ from illustrative response probabilities, compare three ways of
allocating the payoff effect across a model architecture, and show why a direct context-label
implementation need not attain the lower bound.

The question was originally motivated by contextuality and by the Quantum Tug-of-War (QTOW)
line of work \cite{KimQTOW2026,KimQTOW22026}, where a shared state is reused while different
operations generate context-dependent statistics. The present paper abstracts that representational
question away from the specific QTOW dynamics and from quantum ontology. Ontological and
quantum contextuality are considered later as a specialization; in particular, a positive
$I(C;O\mid S)$ is not by itself a Kochen--Specker contextuality witness or contextuality monotone.

The contribution is threefold. First, we formulate a general shared-state framework that separates
state internalization, direct contextual readout, and auxiliary mediation. Second, we derive the
information decomposition and lower bounds for the auxiliary-mediation architecture, including
conditions for saturation. Third, we show how the result can be used in cognitive model analysis
through the recognition-memory example and a practical workflow for comparing alternative
representational allocations.

The remainder of the paper is organized as follows. Section~\ref{sec:framework} introduces the
shared-state framework and information quantities. Section~\ref{sec:obstruction} states and proves
the information-theoretic result and clarifies its scope. Section~\ref{sec:illustration} develops the
recognition-memory worked example. Section~\ref{sec:quantum_relation} gives the ontological
specialization and separates the present CMI from contextuality witnesses. Section~\ref{sec:cognitive_implications}
discusses implications for cognitive model design and empirical analysis, and
Section~\ref{sec:discussion} concludes.

\section{Shared-State Representational Framework}
\label{sec:framework}

The formal result developed below does not require a physical interpretation of the
state variable. We therefore begin at the level of a generic context-sensitive model.
Let $C\in\mathcal{C}$ denote a context or intervention, let $S\in\mathcal{S}$ denote
a candidate internal or latent state representation, and let $O\in\mathcal{O}$ denote
an observable response or outcome. Throughout Sections~\ref{sec:framework}--\ref{sec:illustration},
these variables are treated as discrete. The specialization to ontological models and
quantum contextuality is deferred to Section~\ref{sec:quantum_relation}.

\subsection{Shared-state reuse}

By a \emph{shared-state representation} we mean that the same state variable $S$,
with the same state space $\mathcal{S}$ and semantic interpretation, is reused across
the contexts under comparison. The term does not imply that the state is scalar,
low-dimensional, or context independent in distribution. In particular, no assumption
such as $I(S;C)=0$ is required for the information-theoretic result below. Rather, the
constraint is representational: the model does not replace $\mathcal{S}$ by distinct
context-indexed copies solely in order to encode which context is active.

This distinction matters because context-sensitive behavior can be represented in more
than one way. If the conditional response distribution $p(o\mid s,c)$ varies with $c$
for a fixed candidate state $s$, then the context dependence can be allocated to different
parts of the model architecture.

\subsection{Where context dependence can be represented}
\label{subsec:allocation}

Three broad representational choices will be useful throughout the paper.

\begin{enumerate}
\item \textbf{State internalization.} Context-relevant distinctions can be incorporated
into a richer state representation, for example by replacing $S$ with a refined state
$\widetilde S$ that carries the distinctions needed for context-sensitive responding.
Here $\widetilde S$ is not an additional state carried alongside $S$; it denotes a
redefined state representation that contains the information represented by $S$ together
with the context-relevant distinctions needed for responding.

\item \textbf{Context-dependent response rule.} The state can remain shared while the
response mechanism accesses the context directly, through a conditional law
$p(o\mid s,c)$.

\item \textbf{Auxiliary context representation.} The state can remain shared and the
response mechanism can be made context independent once an auxiliary variable $M$ is
specified. In this case the architecture satisfies
\begin{equation}
 p(o\mid s,m,c)=p(o\mid s,m),
 \label{eq:markov_condition}
\end{equation}
or equivalently
\begin{equation*}
 O\perp C\mid(S,M).
\end{equation*}
The variable $M$ may represent a decision criterion, task set, control state, retrieved
context, routing variable, or other auxiliary information used by the model. It need not
correspond to a literal or anatomically separate memory register.
\end{enumerate}

The result in Section~\ref{sec:obstruction} concerns the third choice. It does not claim
that every context-sensitive model must introduce $M$. Rather, it quantifies a constraint
on models that choose to preserve a shared state representation and to mediate the remaining
context dependence through an auxiliary variable instead of allowing the response mechanism
to access $C$ directly.

\subsection{Information quantities}

All logarithms are base 2 and all information quantities are measured in bits \cite{Shannon1948,CoverThomas}. For a
discrete auxiliary variable $M$, its entropy and its conditional entropy given the candidate state are
\begin{align*}
H(M)&=-\sum_m p(m)\log_2 p(m),\\
H(M\mid S)&=-\sum_s p(s)\sum_m p(m\mid s)\log_2 p(m\mid s).
\end{align*}
The conditional mutual information between context and outcome given the state is
\begin{equation*}
I(C;O\mid S)
=H(O\mid S)-H(O\mid C,S),
\end{equation*}
or, equivalently,
\begin{equation*}
I(C;O\mid S)
=\sum_s p(s)\sum_{c,o}p(c,o\mid s)
\log_2\frac{p(o\mid c,s)}{p(o\mid s)}.
\end{equation*}
Thus $I(C;O\mid S)$ measures the residual statistical dependence between context and
outcome after conditioning on the chosen state representation. It is representation-relative:
its value depends on what distinctions are included in $S$.

The auxiliary-information architecture in Eq.~\eqref{eq:markov_condition} raises a precise
question: if $S$ is kept as the shared state representation and $M$ is used to mediate the
remaining context dependence, how much auxiliary information must $M$ carry beyond what
is already represented by $S$?

\section{Information-Theoretic Bound on Auxiliary Context Representation}
\label{sec:obstruction}

\subsection{Main proposition}

\begin{proposition}[Auxiliary-information bound under shared-state reuse]
\label{thm:info_obstruction}
Let $C,S,M,O$ be discrete random variables with an arbitrary joint distribution satisfying
\begin{equation*}
O\perp C\mid(S,M).
\end{equation*}
Then
\begin{equation}
I(C;M\mid S)
=
I(C;O\mid S)+I(C;M\mid O,S),
\label{eq:exact_mi_decomposition}
\end{equation}
and therefore
\begin{equation}
H(M\mid S)
=
I(C;O\mid S)
+I(C;M\mid O,S)
+H(M\mid C,S).
\label{eq:exact_entropy_decomposition}
\end{equation}
Consequently,
\begin{equation}
I(C;O\mid S)
\le I(C;M\mid S)
\le H(M\mid S)
\le H(M).
\label{eq:bound_chain}
\end{equation}
In particular,
\begin{equation}
H(M\mid S)\ge I(C;O\mid S),
\label{eq:conditional_HM_bound}
\end{equation}
and the unconditional form follows as the corollary
\begin{equation}
H(M)\ge I(C;O\mid S).
\label{eq:HM_bound}
\end{equation}
\end{proposition}

\paragraph*{Proof.}
By the chain rule for conditional mutual information,
\begin{align*}
I(C;M,O\mid S)
&=I(C;M\mid S)+I(C;O\mid M,S)\\
&=I(C;M\mid S),
\end{align*}
where the second equality follows from $O\perp C\mid(S,M)$. Expanding the same
quantity in the opposite order gives
\begin{equation*}
I(C;M,O\mid S)
=I(C;O\mid S)+I(C;M\mid O,S),
\end{equation*}
which proves Eq.~\eqref{eq:exact_mi_decomposition}. Next,
\begin{equation*}
H(M\mid S)=I(C;M\mid S)+H(M\mid C,S).
\end{equation*}
Substitution of Eq.~\eqref{eq:exact_mi_decomposition} yields
Eq.~\eqref{eq:exact_entropy_decomposition}. Since conditional mutual information and
conditional entropy are nonnegative,
\begin{equation*}
I(C;O\mid S)\le I(C;M\mid S)\le H(M\mid S),
\end{equation*}
and conditioning cannot increase entropy, so $H(M\mid S)\le H(M)$. This proves
Eq.~\eqref{eq:bound_chain}. \hfill $\square$

\subsection{Interpretation, tightness, and scope}

The mathematical ingredients of Proposition~\ref{thm:info_obstruction} are standard
Shannon identities. The point of the proposition is representational. For a proposed
shared-state model $S$, the quantity $I(C;O\mid S)$ measures how much context dependence
in the outcome remains after conditioning on that state representation. If the model chooses
to mediate this remaining dependence through an auxiliary variable $M$ satisfying
Eq.~\eqref{eq:markov_condition}, then Eq.~\eqref{eq:exact_mi_decomposition} implies that
$M$ must retain at least $I(C;O\mid S)$ bits of mutual information about context conditional
on $S$. Its conditional entropy $H(M\mid S)$ must consequently be at least as large.

The stronger conditional form in Eq.~\eqref{eq:conditional_HM_bound} is more directly
matched to this interpretation than the unconditional bound in Eq.~\eqref{eq:HM_bound}:
$H(M\mid S)$ excludes the part of the entropy of $M$ that is statistically associated with
$S$ itself. Equation~\eqref{eq:exact_entropy_decomposition} also makes the slack in the
bound explicit. The term $H(M\mid C,S)$ is the residual variability of the auxiliary
representation after context and state are fixed, while $I(C;M\mid O,S)$ is the dependence
between context and the auxiliary representation that remains after conditioning on the
observed outcome and state. These terms are nonnegative, but they should not in general be
interpreted as wasted information: either may reflect legitimate stochastic or latent structure
of a particular cognitive architecture.

The conditional bound is saturated if and only if
\begin{equation*}
H(M\mid C,S)=0
\quad\text{and}\quad
I(C;M\mid O,S)=0.
\end{equation*}
For the unconditional bound $H(M)=I(C;O\mid S)$ to be saturated, one additionally needs
$I(M;S)=0$. Thus tightness is not guaranteed; it depends on the architecture used to mediate
the context dependence. Section~\ref{sec:illustration} gives a simple psychological example
in which the lower bound is strict for the most direct context-label implementation.

The scope of the proposition is specific in three respects. First, Proposition~\ref{thm:info_obstruction} does not
state that a classical representation of context-sensitive behavior is impossible. A model may
instead internalize context-relevant distinctions in a richer state or allow the response rule to
access $C$ directly, as described in Section~\ref{subsec:allocation}. The proposition constrains
only the auxiliary-mediation architecture in Eq.~\eqref{eq:markov_condition}.

Second, $I(C;O\mid S)$ is not an absolute property of a cognitive system. It is relative to
the chosen state representation $S$. Changing the content of the state representation can change
the conditional dependence that remains. This representation dependence is deliberate: the
quantity is intended to diagnose what a proposed state representation leaves for other components
of the model to carry.

Third, positivity of $I(C;O\mid S)$ is not, by itself, a test of Kochen--Specker contextuality
or a contextuality monotone. It indicates residual context dependence relative to $S$. The
relation to ontological contextuality and to witness-based constructions is discussed separately
in Section~\ref{sec:quantum_relation}.

\section{A Psychological Worked Example: Context-Dependent Decision Criteria}
\label{sec:illustration}

To show how Proposition~\ref{thm:info_obstruction} can be used in mathematical
psychology, we consider a standard recognition-memory setting. The purpose of the
example is not to propose a new model of recognition memory. Rather, it illustrates how
the same context-sensitive response statistics can be allocated differently across a cognitive
architecture, and what the information-theoretic bound implies for one such allocation.

\subsection{Recognition judgment with a shared evidence state}

In a conventional old--new recognition task, a participant classifies a test item as
\begin{equation*}
O\in\{\mathrm{old},\mathrm{new}\}.
\end{equation*}
Signal-detection models separate an internal evidence variable from a decision criterion
\cite{MacmillanCreelman2005}. Empirically, payoff and prior-probability manipulations can shift
the cutoff used for old--new responses \cite{HealyKubovy1978}. This makes criterion setting a
useful example of a context-sensitive response mechanism.

For compatibility with the discrete framework used here, let $S$ denote a coarse-grained
candidate memory-evidence state and let $C$ denote the decision context. Because $S$ need not
encode all trial-level variability, the response distribution may remain stochastic even after a
particular state $s$ is fixed.
For concreteness, consider two payoff contexts,
\begin{equation*}
C\in\{L,K\},
\end{equation*}
where $L$ encourages a relatively liberal old response and $K$ encourages a relatively
conservative old response. The same candidate evidence state $s$ can therefore be associated
with different response probabilities,
\begin{equation*}
p(\mathrm{old}\mid s,L)\neq p(\mathrm{old}\mid s,K).
\end{equation*}
The relevant modeling question is not whether such a context effect can be represented
classically---it plainly can---but where the context dependence is placed in the model.

\subsection{Three representational choices}

The alternatives in Section~\ref{subsec:allocation} have a direct interpretation in this task.
A model can (i) incorporate criterion-related distinctions into a richer internal state,
(ii) let the response rule access the payoff context directly, or (iii) preserve the candidate
evidence state $S$ while introducing an auxiliary criterion or control state $M$. The third
architecture satisfies
\begin{equation*}
p(o\mid s,m,c)=p(o\mid s,m),
\end{equation*}
so that the context affects the response only through the auxiliary criterion representation.

Figure~\ref{fig:three_architectures} summarizes these three allocations. The proposition does
not rank them or claim that the third is psychologically correct. It asks a narrower question:
if a model adopts the third architecture, how much context-related information must the
auxiliary mechanism be capable of carrying?

\begin{figure}[t]
\centering
\begin{tikzpicture}[scale=0.88, transform shape,
  node distance=8mm and 11mm,
  box/.style={draw, rounded corners, align=center, minimum width=21mm, minimum height=8mm},
  arr/.style={->, thick}
]

\node[font=\small\bfseries, text width=30mm, align=center] (ta) at (0,2.2) {(A) State\\internalization};
\node[box] (ca) at (-1.2,1.2) {$C$};
\node[box] (s0a) at (-1.2,0) {$S$};
\node[box] (sa) at (1.2,0.6) {$\widetilde S$};
\node[box] (oa) at (1.2,-0.6) {$O$};
\draw[arr] (ca) -- (sa);
\draw[arr] (s0a) -- (sa);
\draw[arr] (sa) -- (oa);

\node[font=\small\bfseries, align=center] (tb) at (5.0,2.2) {(B) Direct\\contextual readout};
\node[box] (cb) at (3.8,1.2) {$C$};
\node[box] (sb) at (6.2,1.2) {$S$};
\node[box] (ob) at (5.0,0) {$O$};
\draw[arr] (cb) -- (ob);
\draw[arr] (sb) -- (ob);

\node[font=\small\bfseries, text width=30mm, align=center] (tc) at (10.1,2.2) {(C) Auxiliary\\criterion};
\node[box] (cc) at (8.8,1.2) {$C$};
\node[box] (mc) at (11.4,1.2) {$M$};
\node[box] (sc) at (8.8,0) {$S$};
\node[box] (oc) at (11.4,0) {$O$};
\draw[arr] (cc) -- (mc);
\draw[arr] (mc) -- (oc);
\draw[arr] (sc) -- (oc);
\node[align=center] at (10.1,-1.05) {$H(M\mid S)\ge I(C;O\mid S)$};

\end{tikzpicture}
\caption{Three ways of allocating context dependence in a cognitive model. In the first,
context-relevant distinctions are incorporated into a richer state $\widetilde S$, which denotes
a refined replacement for $S$ rather than a second state carried alongside it. In the second, the
response rule accesses context directly. In the third, a shared state $S$ is retained and context
dependence is mediated by an auxiliary criterion or control state $M$. Proposition~\ref{thm:info_obstruction}
constrains the third architecture. The arrows indicate the modeled routes by which context may
affect the response; the schematic does not impose statistical independence between $C$ and $S$.}
\label{fig:three_architectures}
\end{figure}

\subsection{Numerical calculation at a fixed evidence state}

Consider one evidence state $S=s_0$ and suppose the two payoff contexts occur with equal
probability,
\begin{equation*}
p(C=L\mid s_0)=p(C=K\mid s_0)=\frac12.
\end{equation*}
The following probabilities are deliberately illustrative rather than estimates from a particular
experiment:
\begin{align*}
p(\mathrm{old}\mid L,s_0)&=0.8,\\
p(\mathrm{old}\mid K,s_0)&=0.2.
\end{align*}
It follows that
\begin{equation*}
p(\mathrm{old}\mid s_0)=\frac12(0.8)+\frac12(0.2)=0.5.
\end{equation*}
Let
\begin{equation*}
h_2(p)=-p\log_2p-(1-p)\log_2(1-p)
\end{equation*}
denote the binary entropy. Then
\begin{equation*}
H(O\mid S=s_0)=h_2(0.5)=1\ \text{bit},
\end{equation*}
whereas
\begin{align*}
H(O\mid C,S=s_0)
&=\frac12 h_2(0.8)+\frac12 h_2(0.2)\\
&=h_2(0.8)\\
&\simeq 0.7219\ \text{bits}.
\end{align*}
Therefore,
\begin{equation}
\boxed{
I(C;O\mid S=s_0)
=1-h_2(0.8)
\simeq 0.2781\ \text{bits}.
}
\label{eq:recognition_cmi}
\end{equation}
For this candidate evidence state, the number in Eq.~\eqref{eq:recognition_cmi} quantifies
the context dependence in the old--new response distribution that remains after $S=s_0$ has
been fixed.

\subsection{Implication for an auxiliary criterion mechanism}

Suppose now that a model preserves the evidence state $S$ and uses an auxiliary criterion
state $M$ to mediate the payoff effect, so that
\begin{equation*}
O\perp C\mid(S,M).
\end{equation*}
Proposition~\ref{thm:info_obstruction} then gives, at $S=s_0$,
\begin{equation}
H(M\mid S=s_0)
\ge I(C;M\mid S=s_0)
\ge I(C;O\mid S=s_0)
\simeq0.2781\ \text{bits}.
\label{eq:recognition_bound}
\end{equation}
Thus an architecture that keeps the candidate evidence representation fixed and does not let
the response mechanism access the payoff context directly must retain at least $0.2781$ bits
of mutual information about context in the auxiliary criterion variable, conditional on the state;
its conditional entropy cannot be smaller than this amount for the illustrative response
distributions above.

This statement should not be read as asserting the existence of a literal $0.2781$-bit memory
module. The bound concerns the information carried by a model variable under a particular
representational decomposition. A concrete psychological implementation may use a richer
criterion state, task set, control process, or other latent mechanism.

\subsection{Why the lower bound need not be tight}

The simplest implementation is to copy the two-valued context into the criterion state,
$M=C$. Because the contexts are equiprobable at $s_0$,
\begin{equation*}
H(M\mid S=s_0)=1\ \text{bit},
\end{equation*}
which is strictly larger than the lower bound in Eq.~\eqref{eq:recognition_bound}. The exact
decomposition in Eq.~\eqref{eq:exact_entropy_decomposition} explains the gap. For $M=C$,
\begin{equation*}
H(M\mid C,S=s_0)=0
\end{equation*}
and
\begin{equation*}
I(C;M\mid O,S=s_0)
=H(C\mid O,S=s_0)
\simeq0.7219\ \text{bits},
\end{equation*}
so that
\begin{equation*}
1=0.2781+0.7219.
\end{equation*}
The latter term is not ``wasted'' information; it reflects the fact that a single stochastic
old--new response does not fully reveal which criterion context was active.

Indeed, for this two-context example the conditional lower bound cannot be saturated. Equality
would require $H(M\mid C,S=s_0)=0$, so $M$ would have to be a deterministic function of
$C$ at the fixed state. Because the two contexts generate different response distributions, a
mediating variable satisfying $O\perp C\mid(S,M)$ cannot map both contexts to the same value
of $M$. It must therefore distinguish $L$ from $K$, which under equal context probabilities gives
$H(M\mid S=s_0)=1$ bit rather than $0.2781$ bits.

\subsection{Interpretation of the worked example}

The example illustrates a use of the bound in cognitive modeling, not a claim that recognition
memory exhibits Kochen--Specker contextuality. Here $I(C;O\mid S)$ is a representation-relative
measure of residual context dependence. It depends on the candidate state $S$ and on the joint
distribution generated by the task design. Nor is the auxiliary variable $M$ uniquely specified by
the theorem. Different cognitive architectures may allocate the same behavioral context effect
to state, response rule, criterion control, or combinations of these components.

The practical role of the bound is therefore diagnostic. Given a proposed shared state, it asks
how much context-sensitive response information remains unabsorbed by that representation and,
for an auxiliary-mediation architecture, gives an information-theoretic floor on what the
additional mechanism must carry.

\section{Relation to Ontological Contextuality and Quantum Theory}
\label{sec:quantum_relation}

The framework in Sections~\ref{sec:framework}--\ref{sec:illustration} was stated without a
physical interpretation of the shared state. This section explains how the same formal structure
specializes to ontological models and how it relates, more cautiously, to contextuality in quantum
theory.

\subsection{Ontological specialization}

For a common preparation in an ontological model, the generic state $S$ can be specialized to
an ontic state $\lambda\in\Lambda$, with preparation distribution $\mu(\lambda)$ and response functions
$\xi(o\mid c,\lambda)$. Operational statistics then take the form
\begin{equation*}
p(o\mid c)
=
\int_{\Lambda}\mu(\lambda)\,\xi(o\mid c,\lambda)\,d\lambda.
\end{equation*}
The shared-state assumption means that the same ontic state space $\Lambda$ is used across
the contexts under comparison rather than being replaced by context-indexed copies solely for
bookkeeping purposes. If the remaining context dependence is mediated through an auxiliary
variable $M$ satisfying
\begin{equation*}
p(o\mid \lambda,m,c)=p(o\mid \lambda,m),
\end{equation*}
then Proposition~\ref{thm:info_obstruction} applies with $S=\lambda$.

This specialization is one motivation for the terminology of shared-state reuse, but it is not
required for the psychological interpretation developed earlier. In particular, the candidate
state $S$ in a cognitive model need not be an ontic state in the sense of quantum foundations.

\subsection{Residual context dependence is not itself a contextuality witness}

The quantity $I(C;O\mid S)$ should not be identified with Kochen--Specker contextuality or
with a contextuality monotone. A positive value says that, relative to the chosen state
representation $S$, the context remains statistically informative about the outcome. This is a
weaker and more general statement than the failure of a noncontextual ontological model.
Contextuality in the latter sense depends on additional structure, such as operational
compatibility or equivalence relations and the requirement that these be represented
noncontextually \cite{KochenSpecker1967,Spekkens2005,AbramskyBrandenburger2011}.
Accordingly, the present paper makes no universal claim of the form that noncontextuality is
equivalent to $I(C;O\mid S)=0$.

Under additional structural restrictions, witness-based information bounds can nevertheless
be constructed. In related work on a restricted external-label simulation class, a distinct
optimized bookkeeping cost was defined and linear witness violations were shown to provide
lower bounds on that cost \cite{KimQTOW22026}. That witness-based quantity is not the same
as the representation-relative conditional mutual information used here. The earlier result is
therefore best viewed as showing that a connection to contextuality witnesses is possible after
the simulation class and zero-obstruction set are specified, rather than as supplying a universal
witness interpretation of $I(C;O\mid S)$.

\subsection{Relation to quantum probability}

Quantum theory represents outcome probabilities by
\begin{equation*}
p(o\mid c)=\mathrm{Tr}\!\left(\rho E_o^{(c)}\right),
\end{equation*}
where $\rho$ is a quantum state and $\{E_o^{(c)}\}$ is the measurement associated with
context $c$. The formalism does not, by itself, posit a single underlying classical random
variable whose context-independent values reproduce all quantum measurement statistics.
No-go results such as Kochen--Specker show that such noncontextual hidden-variable
representations are unavailable for appropriate quantum scenarios
\cite{KochenSpecker1967,Spekkens2005,AbramskyBrandenburger2011}.

The present information-theoretic result should therefore not be described as a theorem that
quantum theory ``evades'' inside the same classical architecture. Rather, the theorem concerns a
particular representational choice: a shared classical state together with an auxiliary variable
that mediates context dependence. Quantum contextuality was historically one motivation for
asking where context-dependent distinctions must reside, and the QTOW framework provided a
concrete decision-dynamical setting in which that question became explicit
\cite{KimQTOW2026,KimQTOW22026}. The bound itself, however, is independent of Hilbert
space, the Born rule, or any claim that cognitive systems are quantum physical systems.

\section{Implications for Context-Sensitive Cognitive Modeling}
\label{sec:cognitive_implications}

The main use of the framework in mathematical psychology is not to classify a behavioral
effect as quantum or nonclassical. It is to make explicit where a model places the information
needed for context-sensitive responding. This question is related to, but different from, the
usual notion of state-space complexity.

\subsection{State-space complexity and contextual representational burden}

A long tradition in mathematical psychology asks how many states or how much internal
structure a model requires in order to reproduce observed behavior. For example, formal work
on Markov learning models has derived lower bounds on the number of states needed to account
for particular learning phenomena \cite{KatsikopoulosFisher2001}, while predictive-state and
partially observable models ask whether a state is sufficient for prediction and control
\cite{Kaelbling1998,Littman2002}.

The quantity $I(C;O\mid S)$ addresses a different question. It does not measure the number
of states, the dimension of $S$, or the number of free parameters. Instead it asks how much
context dependence in the response remains after conditioning on the state representation that
the model has chosen. A large or high-dimensional state can still yield $I(C;O\mid S)>0$ if
context remains informative about the outcome at fixed state. Conversely, a compact state can
in principle give $I(C;O\mid S)=0$ if it already contains all distinctions needed for the
context-sensitive response distribution. The bound therefore concerns the allocation of
behaviorally relevant context information rather than raw state-space size.

This distinction makes the result useful for comparing models that fit the same behavioral
data but distribute contextual information differently. One model may enlarge its latent state,
another may use a context-dependent response rule, and a third may preserve a shared state and
introduce a criterion, task-set, or control variable. Proposition~\ref{thm:info_obstruction}
does not rank these architectures. It supplies a constraint on the third: once $S$ is fixed as the
shared representation, an auxiliary mediator must retain at least $I(C;O\mid S)$ bits of
context information conditional on $S$, and its conditional entropy cannot be smaller than that
amount.

\subsection{A practical model-analysis workflow}

The framework suggests a simple way to use the bound in an applied cognitive-modeling
analysis.

\begin{enumerate}
\item \textbf{Specify the candidate state representation.} The modeler first states what $S$
represents---for example memory evidence, latent value, belief, accumulated evidence, or a
state inferred by a cognitive architecture. This step is essential because the bound is relative
to that representation.

\item \textbf{Characterize context-specific response distributions at fixed state.} From a
fitted model, experimental stratification, or another justified state-estimation procedure, one
obtains $p(o\mid c,s)$ and the corresponding joint distribution needed to compute
$I(C;O\mid S)$. If $S$ is latent, this quantity is model-dependent and is not in general
identified from the marginal behavioral distribution $p(o\mid c)$ alone.

\item \textbf{Locate the context dependence in the architecture.} The modeler asks whether
the contextual effect is intended to be carried by a richer state, by direct access to $C$ in the
response rule, or by an auxiliary variable $M$. Only the last allocation invokes the bound in
Eq.~\eqref{eq:conditional_HM_bound}.

\item \textbf{Compare the implied representational burden.} For an auxiliary-mediation model,
$I(C;O\mid S)$ is an information-theoretic floor on $H(M\mid S)$. The exact decomposition
in Eq.~\eqref{eq:exact_entropy_decomposition} then shows why the realized architecture may
require more conditional entropy than this floor.
\end{enumerate}

This workflow turns the proposition into a diagnostic for model design rather than a generic
measure of behavioral complexity. The recognition-memory calculation in Section~\ref{sec:illustration}
provides the simplest example: after fixing a candidate evidence state, the residual effect of
payoff context can be quantified and then compared with the information carried by a proposed
criterion mechanism.

\subsection{Empirical identification and model comparison}

The framework also clarifies an important identification issue. Behavioral context effects alone
do not determine where contextual information is represented. The same $p(o\mid c)$ may be
compatible with state internalization, direct context-dependent readout, auxiliary control, or
mixtures of these mechanisms. Consequently, $I(C;O\mid S)$ should be interpreted only after
a candidate state variable has been specified or estimated.

This is not a defect of the measure but a statement about the level at which the theory operates.
The bound evaluates the sufficiency of a proposed representational decomposition; it does not
recover that decomposition uniquely from behavior. Empirical discrimination among alternative
architectures therefore requires additional assumptions, manipulations, process measures, or
model-comparison criteria. In this sense the information-theoretic result complements, rather
than replaces, substantive psychological modeling.

\section{Discussion and Outlook}
\label{sec:discussion}

The analysis identifies a representational constraint on context-sensitive models that preserve a
shared state and route the remaining context dependence through an auxiliary variable. Once a
candidate state $S$ is specified, $I(C;O\mid S)$ quantifies the response information about context
that remains unresolved by that state. Proposition~\ref{thm:info_obstruction} shows that an
auxiliary mediator satisfying the conditional-independence architecture must carry at least that
much context information conditional on $S$, and its conditional entropy cannot be smaller.

The result is therefore complementary to conventional measures of model or state-space
complexity. It does not rank state internalization, direct contextual readout, and auxiliary
mediation; rather, it makes their representational commitments explicit. The recognition-memory
example illustrates the point in familiar psychological terms and also shows that the lower bound
can be strict for a natural implementation such as copying the context into a criterion variable.

These considerations also suggest an application to artificial-agent memory design. As agent
architectures increasingly separate active context, persistent memory, retrieval, and control
mechanisms, memory has become a core architectural component rather than an incidental add-on
\cite{ZhangAgentMemory2025}. The present framework offers a way to formulate the resulting
allocation problem without assuming a particular implementation: after a candidate internal state
has been fixed, how much context-sensitive response information remains outside it? A related
recurrent context-switching study operationalizes this distinction in one concrete setting
\cite{KimContextualControl2026}; broader agent architectures remain to be studied.

Two qualifications are central to the interpretation. First, the diagnostic is relative to the chosen
state representation: changing $S$ can change the residual dependence $I(C;O\mid S)$. Second,
that residual dependence is not itself a test of Kochen--Specker contextuality. The ontological
specialization in Section~\ref{sec:quantum_relation} explains the connection that motivated the
problem without making quantum structure a prerequisite for the cognitive application.

Several extensions remain open. Continuous-state formulations require measure-theoretic care;
dynamical models raise questions about how the representational burden evolves over time;
approximate simulation would require a controlled notion of response-distribution error; and
empirical work must address estimation of $I(C;O\mid S)$ when $S$ is latent. Restricted
simulation classes may also support stronger witness-based information bounds, as in the related
external-bookkeeping analysis discussed in Section~\ref{sec:quantum_relation}.

The central question is therefore not whether context-sensitive behavior is classical or quantum,
but how a model allocates the information required for that behavior. The framework provides a
compact way to expose that allocation once a candidate shared state has been specified, and the
resulting bound supplies a quantitative constraint on architectures that assign the remaining
context dependence to an auxiliary mechanism.

\section*{Acknowledgements}
This work was supported by SOBIN Institute LLC under Research Grant SP005.

\section*{Declaration of Generative AI and AI-assisted technologies in the writing process}
During the preparation of this work the author used ChatGPT (OpenAI) for English editing and manuscript preparation support. After using this tool, the author reviewed and edited the content as needed and takes full responsibility for the content of the publication.

\end{document}